# Text-Based Correlation Matrix in Multi-Asset Allocation

Yasuhiro Nakayama [1], Tomochika Sawaki [2], Issei Furuya [2], Shunsuke Tamura [1]

[1] Mizuho Research & Technologies, Ltd.
[2] Mizuho Bank, Ltd.

**Abstract:** The purpose of this study is to estimate the correlation structure between multiple assets using financial text analysis. In recent years, as the background of elevating inflation in the global economy and monetary policy tightening by central banks, the correlation structure between assets, especially interest rate sensitivity and inflation sensitivity, has changed dramatically, increasing the impact on the performance of investors' portfolios. Therefore, the importance of estimating a robust correlation structure in portfolio management has increased. On the other hand, the correlation coefficient using only the historical price data observed in the financial market is accompanied by a certain degree of time lag, and also has the aspect that prediction errors can occur due to the nonstationarity of financial time series data, and that the interpretability from the viewpoint of fundamentals is a little poor when a phase change occurs. In this study, we performed natural language processing on news text and central bank text to verify the prediction accuracy of future correlation coefficient changes. As a result, it was suggested that this method is useful in comparison with the prediction from ordinary time series data.

## 1 Introduction

In recent years, the correlation structure between assets has changed significantly against the background of sharp increases in inflation and changes in monetary policy of central banks in various countries. Figure 1 shows data on the S & P500 index of U.S. equities and the USD-denominated Total Return Index of U.S. Government Bonds obtained from Bloomberg from January 1973 to January 2024. Monthly returns are calculated, and the correlation coefficients over the past 24 months are rolled out over time. Although it should be noted that this is a simple analysis that does not use the total return index including stock dividends because of the relationship between the period of data acquisition, it is considered that in the past 20 years, on average, there have been many negative correlation coefficients, i.e., markets in which either safe assets or risky assets are bought, which are called risk-on or risk-off when calculated on a monthly basis, while there have been many positive correlation coefficients, i.e., markets in which price fluctuations of safe assets and risky assets are in the same direction, since 2021.

Since changes in the correlation structure between assets are a serious problem in portfolio management, the importance of estimating changes in the correlation structure is increasing. On the other hand, the correlation structure estimated from historical price data involves time lags, so it is difficult to deal with the situation when a change occurs. In this study, we aim to generate inflation and economic growth scores from news and central bank texts and use them to estimate correlation structures corresponding to changes in macroeconomic conditions.

Figure 1: Correlation between stocks and bonds in the United States

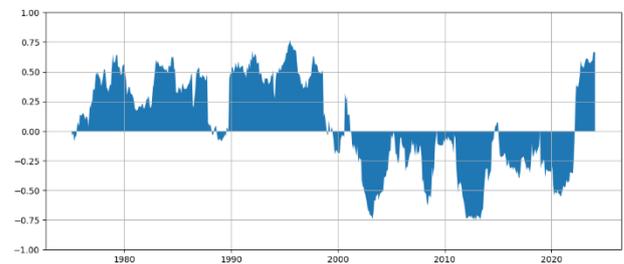

## 2 Related research

There are many studies on the correlation between stocks and bonds using the correlation coefficient during declines. Ilmanen focused on the variables of the correlation and showed that economic growth prospects and inflation affected the correlation [1]. Brixton et al. also analyzed the long-term correlation between U.S. stocks and bonds and concluded that they affects the volatility and correlation between economic growth and inflation, not the level of inflation [2].

Efforts are also being made to calculate scores linked to economic indicators such as inflation and economic growth from text. Nakayama and Sawaki [3] calculated scores for inflation and economic growth using an implication judgment model for documents published by the Fed.

Many macroeconomic indicators, like CPI and GDP, are released on a monthly or quarterly basis, which makes it challenging to assess inflation and economic growth promptly. Hence, in this study, we computed inflation and economic growth scores from daily news texts, attempting to understand their situation in a more timely manner. The calculated scores were used to predict changes in the correlation coefficient, to verify the utility of these scores, and to analyze the factors influencing changes in the correlation structure.

## 3 Text-Based Score

### 3.1 Data and Preprocessing

We utilized Bloomberg news data spanning from January 2009 to November 2023, restricting the analysis to English-language content only. Furthermore, we applied the topic codes assigned to the news to narrow our focus to articles about the United States, aiming to exclude headlines that merely mentioned rising or falling stock prices as much as possible. Additionally, for news circulated multiple times, for instance, through added text, only the initial instance was considered. Following the aforementioned screening process, we analyzed approximately 20 million news items over a period of about 15 years.

### 3.2 Analytical Methods

We analyze two topics: inflation and economic growth. First, we use a zero-shot-based implication judgment mode[1]l to determine whether each headline implies the meaning of a hypothetical sentence. The hypothesis statement was as follows.

(Inflation)

Ascending Hypothesis Statement: "Inflation rate will increase."

Descending Hypothesis Statement: "Inflation rate will decline."

(Economic Growth)

Ascending Hypothesis Statement: "Economic growth will increase."

Descending Hypothesis Statement: "Economic growth will decline."

We did not learn the entailment model, but used the publicly available learned model. The entailment model returns a score of 0~1 whether the judgment target sentence implies the meaning of the hypothesis sentence. A score closer to 1 implies a hypothetical statement; The closer it is to 0, the more it contradicts the hypothesis statement. In this study, it was judged that the score (score of the ascending hypothesis sentence - score of the descending hypothesis sentence) implied the ascending hypothesis sentence when it exceeded 0.8, and that it implied the descending hypothesis sentence when it fell below -0.8.

### 3.3 Calculation of the score

Scores were tallied on a weekly basis. The news published from Monday to Sunday each week was targeted, and the number of news items that exceeded a certain threshold was counted for each topic - inflation and economic growth. They were then converted into a score ranging from -1 to 1 for each topic using the following formula.

$$Score_{topic}(t) = (C_{up}(t) - C_{down}(t))/(C_{up}(t) + C_{down}(t))$$

where is the number of news items judged to imply a rising hypothesis statement at the time point, and is the number of news items judged to imply a falling hypothesis statement at the time point. $C_{up}(t) t C_{down}(t) t$

## 4 Prediction of correlation coefficient

### 4.1 Problem Setting

(Data)

The forecast is based on the difference between the correlation coefficients of stocks and bonds over the forthcoming 6 months (125 business days) and the past 6 months. The analysis included three countries and regions: the United States, the United Kingdom, and Europe. For stocks, the dividend index was used, and for bonds, the total return index was utilized. All data were sourced from Bloomberg, with the daily rate of change being calculated. Additionally, forecasts were generated on a weekly basis.

(Learning Period and Evaluation Period)

The evaluation period spanned from April 2019 to August 2023. The study period commenced in August 2009, and was revisited annually using data available at the start of each year. However, any timeframe requiring teacher (or 'label') data from within the evaluation period was excluded from the dataset to avoid data leakage.

### 4.2 Proposed Model

We constructed a forecasting model utilizing features based on news scores (inflation and economic growth) developed in Chapter 3. XGBoost was employed as the machine learning model. The features used are as follows.

---
[1] https://huggingface.co/facebook/bart-large-mnli

(List of Features)
・New Text Score (Inflation and Economic Growth)
・Difference from the average score over the past 12 weeks
・Correlation of score variability over the past 12 weeks
・Ratio of change in score over the past 12 weeks
・Fed Fund rate
・difference from three months before FF rate[2]

(Features used only for the United States)
・Score (inflation/economy) calculated from Minutes (created using the same method as in [3])
・Difference in Minutes score from 3 months ago

4.3 Results

Table 1 shows the RMSE for the evaluation period. In all countries and regions, the RMSE of the proposed model was lower than that of the two benchmark models. These results suggest the usefulness of text scores for predicting correlation changes.

Table 1: RMSE for the evaluation period

|    | BM1  | BM2  | proposed model |
|----|------|------|----------------|
| US | 0.49 | 0.30 | **0.25**       |
| UK | 0.46 | 0.27 | **0.21**       |
| EU | 0.65 | 0.37 | **0.32**       |

*BM1: This benchmark model, the correlation coefficient, calculated using data from 12 to 6 months ago, and the change in the correlation coefficient over the last 6 months, are presented side by side.

*BM2: This benchmark model featuring a zero change in the correlation coefficient

## 5 Feature analysis

Next, we will analyze the features contributing to the prediction of changes in correlations. We will focus on the in-sample learning results, utilizing all available data as the learning dataset, instead of considering out-of-sample results. Specifically, to examine the characteristics of scores derived from news data, we exclusively used the news scores (for inflation and economic growth) generated in Chapter 3. This included the deviation of each score from its 12-week average, and the change in scores from the previous week, as feature values. The analysis targeted the United States population.

Figure 2 shows the average of absolute SHAP values for each feature value. The most significant contribution was "the ratio of the inflation score and the economic growth score over the past 12 weeks, " followed by " the economic growth score, " "the inflation score, " and " the correlation between the inflation score and the economic growth score over the past 12 weeks. "

Figure 2: Absolute mean SHAP values for each feature

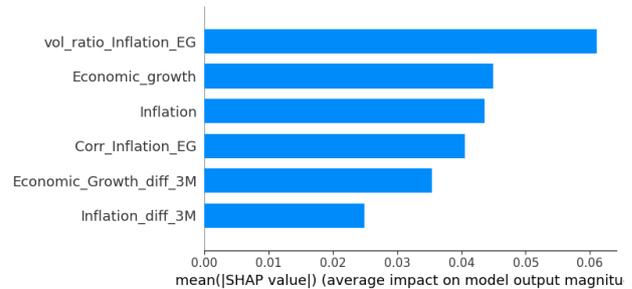

* vol_ratio_inflation_EG represents the ratio of the changes in the inflation and economic growth scores over the last 12 weeks, as derived from news sources. Corr_Inflation_EG measures the correlation between the changes in the inflation and economic growth scores over the same period, also drawn from news sources.

Figures 3 and 4 display the SHAP values for the "ratio of the rate of change of the inflation score to its 12-week historical average" and the "economic growth score," correspondingly. Examining the SHAP value for the ratio of change rate in the inflation score depicted in Figure 3, it becomes apparent that the positive correlation escalates when the ratio is substantial (namely, when economic growth remains stable, whereas the inflation outlook is uncertain). Observing the SHAP value for the "economic growth score" in Figure 4, one notes that the positive correlation heightens with a higher economic growth score. This finding correlates with our qualitative analysis that economic growth benefits stocks yet adversely affects bonds.

---

[2] FF rates were sourced from FRED®. https://fred.stlouisfed.org/series/FEDFUNDS

Figure 3: SHAP value for "Ratio of percent change in score over the past 12 weeks"

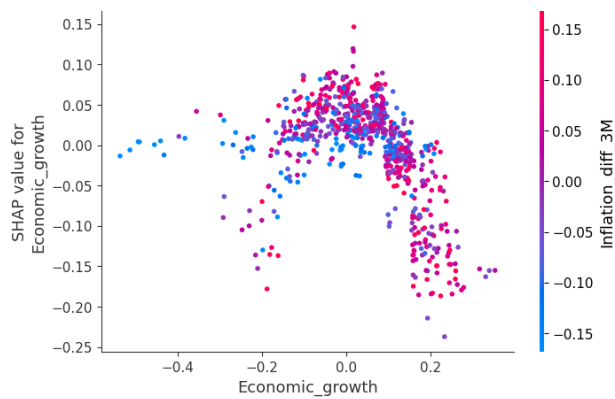

Figure 4: SHAP values for the Economic Growth Score

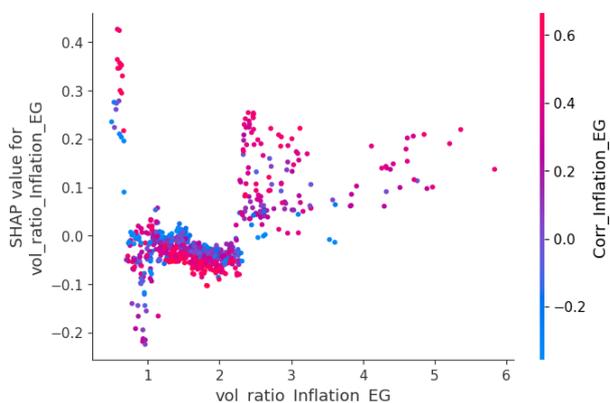

## 6 Summary and Future Issues

In this study, we derived inflation and economic growth scores from news headlines. We then utilized these scores to forecast changes in the correlation coefficients among stocks and bonds in the United States, the United Kingdom, and Europe. Our findings indicate that the scores significantly contributed to the predictions of correlation changes.

Using the SHAP values, we examined how each feature value affected the correlation change prediction.

In the future, we would like to analyze the correlation between different assets so that it can be used to construct multi-asset portfolios.

## Notes

The contents and views of this paper belong to the authors and are not the official views of their companies.

## Acknowledgment

This study expands on the analysis at the BQuant Hackathon. We would like to thank the members of the hackathon for their helpful advice.